# Value Function Approximation in Zero-Sum Markov Games

Michail G. Lagoudakis and Ronald Parr
Department of Computer Science
Duke University
Durham, NC 27708
{mgl,parr}@cs.duke.edu

## Abstract

This paper investigates value function approximation in the context of zero-sum Markov games, which can be viewed as a generalization of the Markov decision process (MDP) framework to the two-agent case. We generalize error bounds from MDPs to Markov games and describe generalizations of reinforcement learning algorithms to Markov games. We present a generalization of the optimal stopping problem to a two-player simultaneous move Markov game. For this special problem, we provide stronger bounds and can guarantee convergence for LSTD and temporal difference learning with linear value function approximation. We demonstrate the viability of value function approximation for Markov games by using the Least squares policy iteration (LSPI) algorithm to learn good policies for a soccer domain and a flow control problem.

## 1 Introduction

Markov games can be viewed as generalizations of both classical game theory and the Markov decision process (MDP) framework[1]. In this paper, we consider the two-player zero-sum case, in which two players make simultaneous decisions in the same environment with shared state information. The reward function and the state transition probabilities depend on the current state and the current agents' joint actions. The reward function in each state is the payoff matrix of a zero-sum game.

Many practical problem can be viewed as Markov games, including military-style pursuit-evasion problems. Moreover, many problems that are solved as classical games by working in strategy space may be represented more effectively as Markov games. A well-known example of a Markov game is Littman's soccer domain (Littman, 1994).

Recent work on learning in games has emphasized accelerating learning and exploiting opponent suboptimalities (Bowling & Veloso, 2001). The problem of learning in games with very large state spaces using function approximation has not been a major focus, possibly due to the perception that a prohibitively large number of linear programs would need to be solved during learning.

This paper contributes to the theory and practice of learning in Markov games. On the theoretical side, we generalize some standard bounds on MDP performance with approximate value functions to Markov games. When presented in their fully general form, these turn out to be straightforward applications of known properties of contraction mappings to Markov games.

We also generalize the optimal stopping problem to the Markov game case. *Optimal stopping* is a special case of an MDP in which states have only two actions: *continue* on the current Markov chain, or *exit* and receive a (possibly state dependent) reward. Optimal stopping is of interest because it can be used to model the decision problem faced by the holder of stock options (and is purportedly modeled as such by financial institutions). Moreover, linear value function approximation algorithms can be used for these problems with guaranteed convergence and more compelling error bounds than those obtainable for generic MDPs. We extend these results to a Markov game we call the *"opt-out"*, in which two players must decide whether to remain in the game or opt out, with a payoff that is dependent upon the actions of both players. For example, this can be used to model the decision problem faced by two "partners" who can choose to continue participating in a business venture, or force a sale and redistribution of the business assets.

As in the case of MDPs, these theoretical results are somewhat more motivational than practical. They provide reassurance that value function approximation architectures which can represent value functions with reasonable errors will lead to good policies. However, as in the MDP case, they do not provide *a priori* guarantees of good performance. Therefore we demonstrate, on the practical side of this paper, that with a good value function approximation architecture, we can achieve good performance on large Markov games. We achieve this by extending the least squares policy iteration (LSPI) algorithm (Lagoudakis & Parr, 2001), originally designed for MDPs, to general two-player zero-sum Markov games. LSPI is an approximate policy iteration algorithm that uses linear approximation architectures, makes efficient use of training samples, and learns faster than conventional learning methods. The advantage of using LSPI for Markov games stems from its generalization ability and its data efficiency. The latter is of even greater importance for large Markov Games than for MDPs since each learning step is not light weight; it requires solving a linear program.

---
[1] Littman (1994) notes that the Markov game framework actually preceded the MDP framework.

We demonstrate LSPI with value function approximation in two domains: Littman's soccer domain in which LSPI learns very good policies using a relatively small number of sample "random games" for training; and, a simple server/router flow control problem where the optimal policy is found easily with a small number of training samples.

## 2 Markov Games

A two-player zero-sum Markov game is defined as a 6-tuple $(\mathcal{S}, \mathcal{A}, \mathcal{O}, P, \mathcal{R}, \gamma)$, where: $\mathcal{S} = \{s_1, s_2, ..., s_n\}$ is a finite set of game states; $\mathcal{A} = \{a_1, a_2, ..., a_m\}$ and $\mathcal{O} = \{o_1, o_2, ..., o_l\}$ are finite sets of actions, one for each player; $P$ is a Markovian state transition model — $P(s, a, o, s')$ is the probability that $s'$ will be the next state of the game when the players take actions $a$ and $o$ respectively in state $s$; $\mathcal{R}$ is a reward (or cost) function — $\mathcal{R}(s, a, o)$ is the expected one-step reward for taking actions $a$ and $o$ in state $s$; and, $\gamma \in (0, 1]$ is the discount factor for future rewards. We will refer to the first player as the "agent," and the second player as the "opponent." Note that if the opponent is permitted only a single action, the Markov game becomes an MDP.

A *policy* $\pi$ for a player in a Markov game is a mapping, $\pi : \mathcal{S} \to \Omega(\mathcal{A})$, which yields probability distributions over the agent's actions for each state in $\mathcal{S}$. Unlike MDPs, the optimal policy for a Markov game may be stochastic, i.e., it may define a *mixed* strategy for every state. By convention, $\pi(s)$ denotes the probability distribution over actions in state $s$ and $\pi(s, a)$ denotes the probability of action $a$ in state $s$.

The agent is interested in maximizing its expected, discounted return in the *minimax* sense, that is, assuming the worst case of an optimal opponent. Since the underlying rewards are zero-sum, it is sufficient to view the opponent as acting to minimize the agent's return. Thus, the state value function for a zero-sum Markov game can be defined in a manner analogous to the Bellman equation for MDPs:

$$V(s) = \max_{\pi(s) \in \Omega(\mathcal{A})} \min_{o \in \mathcal{O}} \sum_{a \in \mathcal{A}} Q(s, a, o) \pi(s, a) \ , \quad (1)$$

where the $Q$-values familiar to MDPs are now defined over states and pairs of agent and opponent actions:

$$Q(s, a, o) = \mathcal{R}(s, a, o) + \gamma \sum_{s' \in \mathcal{S}} P(s, a, o, s') V(s') \ . \quad (2)$$

We will refer to the policy chosen by Eq. (1) as the *minimax policy* with respect to $Q$. This policy can be determined in any state $s$ by solving the following linear program:

Maximize: $V(s)$
Subject to: $\sum_{a \in \mathcal{A}} \pi(s, a) = 1$
$\forall a \in \mathcal{A}, \ \pi(s, a) \geq 0$
$\forall o \in \mathcal{O}, \ V(s) \leq \sum_{a \in \mathcal{A}} Q(s, a, o) \pi(s, a)$

In many ways, it is useful to think of the above linear program as a generalization of the max operator for MDPs (this is also the view taken by Szepesvari and Littman (1999)). We briefly summarize some of the properties of operators for Markov games which are analogous to operators for MDPs. (See Bertsekas and Tsitsiklis (1996) for an overview.) Eq. (1) and Eq. (2) have fixed points $V^*$ and $Q^*$, respectively, for which the minimax policy is optimal in the minimax sense. The following iteration, which we name with operator $\mathcal{T}^*$, converges to $Q^*$:

$$\mathcal{T}^* : Q^{(i+1)}(s, a, o) = \mathcal{R}(s, a, o) + \\ \gamma \sum_{s' \in \mathcal{S}} P(s, a, o, s') \max_{\pi(s) \in \Omega(\mathcal{A})} \min_{o \in \mathcal{O}} \sum_{a \in \mathcal{A}} Q^{(i)}(s, a, o) \pi(s, a) \ .$$

This is analogous to *value iteration* in the MDP case.

For any policy $\pi$, we can define $Q_\pi(s, a, o)$ as the expected, discounted reward when following policy $\pi$ after taking actions $a$ and $o$ for the first step. The corresponding fixed point equation for $Q_\pi$ and operator $\mathcal{T}^\pi$ are:

$$\mathcal{T}^\pi : Q^{(i+1)}(s, a, o) = \mathcal{R}(s, a, o) + \\ \gamma \sum_{s' \in \mathcal{S}} P(s, a, o, s') \min_{o \in \mathcal{O}} \sum_{a \in \mathcal{A}} Q^{(i)}(s, a, o) \pi(s, a) \ .$$

A *policy iteration* algorithm can be implemented for Markov games in a manner analogous to policy iteration for MDPs by fixing a policy $\pi_i$, solving for $Q_{\pi_i}$, choosing $\pi_{i+1}$ as the minimax policy with respect to $Q_{\pi_i}$ and iterating. This algorithm will also converge to $Q^*$.

## 3 Value Function Approximation

Value function approximation has played an important role in extending the classical MDP framework to solve practical problems. While we are not yet able to provide strong *a priori* guarantees in most cases for the performance of specific value function architectures on specific problems, careful analyses such as (Bertsekas & Tsitsiklis, 1996) have legitimized the use of value function approximation for MDPs by providing loose guarantees that good value functions approximations will result in good policies.

Bertsekas and Tsitsiklis (1996) note that "the Neuro-Dynamic Programming (NDP) methodology with cost-to-go function approximation is not very well developed at present for dynamic games"[2] (page 416). To our knowledge, this section of the paper provides the first error bounds on the use of approximate value functions for Markov games. We achieve these results by first stating extremely general results about contraction mappings and then showing that known error bound results for MDPs generalize to Markov games as a result.[3].

An operator $\mathcal{T}$ is a *contraction* with rate $\alpha$ with respect to some norm $\|\cdot\|$ if there exists a real number $\alpha \in [0,1)$ such that for all $V_1$ and $V_2$:

$$\|\mathcal{T}V_1 - \mathcal{T}V_2\| \leq \alpha \|V_1 - V_2\| \ .$$

An operator $\mathcal{T}$ is a *non-expansion* if

$$\|\mathcal{T}V_1 - \mathcal{T}V_2\| \leq \|V_1 - V_2\| \ .$$

The *residual* of a vector $V$ with respect to some operator $\mathcal{T}$ and norm $\|\cdot\|$ is defined as

$$\epsilon_\mathcal{T}(V) = \|\mathcal{T}V - V\| \ .$$

**Lemma 3.1** *Let $\mathcal{T}$ be a contraction mapping in $\|\cdot\|$ with rate $\alpha$ and fixed point $V^*$. Then,*

$$\|V^* - V\| \leq \frac{\epsilon_\mathcal{T}(V)}{1-\alpha} \ .$$

This lemma is verified easily using the triangle inequality and contraction property of $\mathcal{T}$.

**Lemma 3.2** *Let $\mathcal{T}_1$ and $\mathcal{T}_2$ be contraction mappings in $\|\cdot\|$ with rates $\alpha_1$, $\alpha_2$ (respectively) and fixed points $V_1^*$ and $V_2^*$ (respectively). Then,*

$$\|V_1^* - V_2^*\| \leq \frac{\epsilon_{\mathcal{T}_1}(V) + \epsilon_{\mathcal{T}_2}(V)}{1-\alpha_3} \ ,$$

*where $\alpha_3 = \max(\alpha_1, \alpha_2)$.*

This result follows immediately from the triangle inequality and the previous lemma. A convenient and relatively well-known property of Markov games is that operators $\mathcal{T}^*$ and $\mathcal{T}^\pi$ are contractions in $\|\cdot\|_\infty$ with rate $\gamma$, where $\|V\|_\infty = \max_{s \in \mathcal{S}} |V(s)|$ (Bertsekas & Tsitsiklis, 1996). The convergence of value iteration for both Markov games and MDPs follows immediately from this result.

---

[2]NDP is their term for value function approximation.

[3]We thank Michael Littman for pointing out that these results should also apply to all generalized MDPs (Szepesvari & Littman, 1999)

**Theorem 3.3** *Let $Q^*$ be the optimal $Q$ function for a Markov game with value iteration operator $\mathcal{T}^*$.*

$$\|Q^* - Q\| \leq \frac{\epsilon_{\mathcal{T}^*}(Q)}{1-\gamma} \ .$$

This result follows directly from Lemma 3.1 and the contraction property of $\mathcal{T}^*$. While it gives us a loose bound on how far we are from the true value function, it doesn't answer the more practical question of how the policy corresponding to a suboptimal value function performs in comparison to the optimal policy. We now generalize results from Williams and Baird (1993) to achieve the following bound:

**Theorem 3.4** *Let $Q^*$ be the optimal $Q$ function for a Markov game and $\mathcal{T}^*$ be the value iteration operator for this game. If $\pi$ is the minimax policy with respect to $Q$, then*

$$\|Q^* - Q_\pi\| \leq \frac{2\epsilon_{\mathcal{T}^*}(Q)}{1-\gamma} \ .$$

This result follows directly from Lemma 3.2 by noting that $\epsilon_{\mathcal{T}^*}$ is identical to $\epsilon_{\mathcal{T}^\pi}$ at this point. As noted by Baird and Williams the result can be tightened by a factor of $\gamma$ by observing that for the next iteration, $\mathcal{T}^*$ and $\mathcal{T}^\pi$ are identical. Since this bound is tight for MDPs, it is necessarily tight for Markov games.

The significance of these results is more motivational than practical. As in the MDP case, the impact of small approximation errors can, in the worst case, be quite large. Moreover, it is essentially impossible in practice to compute the residual for large state spaces and, therefore, impossible to compute a bound for an actual $Q$ function. As with MDPs, however, these results can motivate the use of value function approximation techniques.

## 4 Approximation in Markov Game Algorithms

Littman (1994) studied Markov games as a framework for multiagent reinforcement learning (RL) by extending tabular $Q$-learning to a variant called minimax-$Q$. By viewing the maximization in Eq. (1) as a generalized max operation (Szepesvari & Littman, 1999) minimax-$Q$ simply uses Eq. (1) in place of the traditional max operator in $Q$-learning. The rub is that minimax-$Q$ must therefore solve one linear program for every experience sampled from the environment. We note that in the case of Littman's $4 \times 5$ soccer domain, $10^6$ learning steps were used[4], which im-plies that $10^6$ linear programs were solved. Even though the linear programs were small, this adds further motivation to the desire to increase learning efficiency through the use of value function approximation.

---

[4]It's not clear that this many were needed; but this was the duration the training period.

In the natural formulation of $Q$-learning with value function approximation for MDPs, a tabular $Q$-function is replaced by a parameterized $Q$-function, $\widehat{Q}(s, a, w)$, where $w$ is some set of parameters describing our approximation architecture, e.g., a neural network or a polynomial. The update equation for a transition from state $s$ to state $s'$ with action $a$ and reward $r$ then becomes:

$$w \leftarrow w + \alpha \left( r + \gamma \max_{a' \in \mathcal{A}} \widehat{Q}(s', a', w) - \widehat{Q}(s, a, w) \right) \nabla_w \widehat{Q}(s, a, w) ,$$

where $\alpha$ is the learning rate.

One could easily extend this approximate $Q$-learning algorithm to Markov games by defining the $Q$-functions over states and agent-opponent action pairs, and replacing the max operator in the update equation with the minimax operator of Eq. (1). This approach would yield the same stability and performance guarantees of ordinary $Q$-learning with function approximation, that is, essentially none. Also, it would require solving a very large number of linear programs to learn a good policy. In the following two sections, we separately address both of these concerns: stable approximation and efficient learning.

## 5 Stable Approximation

A standard approach to proving the stability of a function approximation scheme is to exploit the contraction properties of the update operator $\mathcal{T}$. If a function approximation step can be shown to be non-expansive in the same norm in which $\mathcal{T}$ is a contraction, then the combined update/approximation step remains a contraction. This approach is used by Gordon (1995), who shows that a general approximation architecture called an *averager* can be used for stable value function approximation in MDPs. This and similar results generalize immediately to the Markov game case and may prove useful for continuous space Markov games, such as pursuit-evasion problems.

In our limited space, we choose to focus on an architecture for which the analysis is somewhat trickier: linear approximation architectures. Evaluation of a fixed policy for an MDP is known to converge when the value function is approximated by a linear architecture (Van Roy, 1998). Two such stable approaches are least-squares temporal difference (LSTD) learning (Bradtke & Barto, 1996) and linear temporal difference (TD) learning (Sutton, 1988). In these methods the value function $V$ is approximated by a linear combination of basis functions $\phi_i$:

$$\widehat{V}(s, w) = \sum_{i=1}^{k} \phi_i(s) w_i = \phi(s)^\mathsf{T} w .$$

We briefly review the core of the convergence results and error bounds for linear value function approximation and the optimal stopping problem. We can view the approximation as alternating between dynamic programming steps with operator $\mathcal{T}$ and projection steps with operator $\Pi$, in an attempt to find the fixed point $\widehat{V}$ (and the corresponding $w$):

$$\widehat{V} = \Pi \mathcal{T} \widehat{V} .$$

If the operator $\mathcal{T}$ is a contraction and $\Pi$ is a non-expansion in the same norm in which $\mathcal{T}$ is a contraction, then $w$ is well-defined and iterative methods such as value iteration or temporal difference learning will converge to $w$. For linear value function approximation, if the transition model $P$ yields a mixing process with stationary distribution $\rho$, $\mathcal{T}$ is a contraction in the weighted $L_2$ norm $\| \cdot \|_\rho$ (Van Roy, 1998), where

$$\|V\|_\rho = \left( \sum_{s \in \mathcal{S}} \rho(s) V(s)^2 \right)^{1/2} .$$

If $\Pi_\rho$ is an orthogonal projection weighted by $\rho$, then $w$ is well-defined, since a projection in any weighted $L_2$ norm is necessarily non-expansive in that norm. Using matrix and vector notation, we have

$$\Phi = \begin{pmatrix} \phi(s_1)^\mathsf{T} \\ \ldots \\ \phi(s)^\mathsf{T} \\ \ldots \\ \phi(s_{|\mathcal{S}|})^\mathsf{T} \end{pmatrix} ,$$

and $w$ can be defined in closed form as the result of the linear Bellman operator combined with orthogonal projection:

$$w = (\Phi^\mathsf{T}(\Phi - \gamma \mathbf{P} \Phi))^{-1} \Phi^\mathsf{T} \mathcal{R} .$$

The LSTD algorithm solves directly for $w$, while linear TD finds $w$ by stochastic approximation.

Since $w$ is found by orthogonal projection, the Pythagorean theorem can be used to bound the distance from $\widehat{V}$ to the true value function $V^*$ (Van Roy, 1998):

$$\|V^* - \widehat{V}\|_\rho \leq \frac{1}{\sqrt{1 - \gamma^2}} \|V^* - \Pi_\rho V^*\|_\rho .$$

In general, these algorithms are not guaranteed to work for optimization problems. Linear TD can diverge if a max operator is introduced and LSTD is not suitable for the policy evaluation stage of a policy iteration algorithm (Koller & Parr, 2000). However, these algorithms can be applied to the problem of optimal stopping and an alternating game version of the optimal stopping problem (Van Roy, 1998).

Optimal stopping can be used to model the problem of when to sell an asset or how to price an option. The system evolves as an *uncontrolled* Markov chain because the actions of an individual agent cannot affect the evolution of the system. This characterization is true of individual investors in stock or commodity markets. The only actions are *stopping* actions, which correspond to jumping off the Markov chain and receiving a reward which is a function of the state. For example, selling a commodity is equivalent to jumping off the Markov chain and receiving a reward equivalent to the current price of the commodity.

We briefly review a proof of convergence for linear function approximation for the optimal stopping problem. Let the operator $T_c$ be the dynamic programming operator for the continuation of the process:

$$T_c : V^{(t+1)}(s) = \mathcal{R}_c(s) + \gamma \sum_{s' \in \mathcal{S}} P(s'|s) V^{(t)}(s') \ ,$$

where $\mathcal{R}_c$ is the reward function associated with continuing the process. We can define the stopping operator to be:

$$T_h : V^{(t+1)}(s) = \max(V^{(t)}(s), \mathcal{R}_h(s)) \ ,$$

where $\mathcal{R}_h$ is the reward associated with stopping at state $s$. The form of value iteration for the optimal stopping problem with linear value function approximation is thus: $V^{(t+1)} = \Pi_\rho T_h T_c V^{(t)}$ and the combined operation will have a well-defined fixed point if $T_h$ is non-expansive in $\| \cdot \|_\rho$.

**Lemma 5.1** *If an operator $T$ is pointwise non-expansive then it is non-expansive in any weigthed $L_2$ norm.*

It is easy to see that the maximum operation in the optimal stopping problem has this property: $|\max(c, V_1(s)) - \max(c, V_2(s))| \leq |V_1(s) - V_2(s)|$, which implies $\|T_h V_1 - T_h V_2\|_\rho \leq \|V_1 - V_2\|_\rho$ for any $\rho$ weighted $L_2$ norm.

We now present a generalization of the optimal stopping problem to Markov games. We call this the "opt-out" problem, in which two players must decide whether to continue along a Markov chain with Bellman operator $T_c$ and reward $\mathcal{R}_c$, or adopt one of several possible exiting actions $a_1 \ldots a_i$, and $o_1 \ldots o_j$. For each combination of agent and opponent actions, the game will terminate with probability $P_h(s, a, o)$ and continue with probability $1 - P_h(s, a, o)$. If the game terminates, the agent will receive reward $\mathcal{R}_h(s, a, o)$ and the opponent will receive reward $-\mathcal{R}_h(s, a, o)$. Otherwise, the game continues. This type of game can be used to model the actions of two partners with buy-out options. If both choose to exercise their buy-out options at the same time, a sale and redistribution of the assets may be forced. If only one partner chooses to exercise a buy-out option, the assets will be distributed in a potentially uneven way, depending upon the terms of the contract and the type of buy-out action exercised. Thus, the decision to exercise a buy-out option will depend upon the actions of the other partner, and the expected, discounted value of future buy-out scenarios.[5]

The operator $T_h$ is then defined as the follows:

$$T_h : V^{(t+1)}(s) = \max_\pi \min_{o \in \mathcal{O}} \sum_{a \in \mathcal{A}} \Big( P_h(s, a, o) \mathcal{R}_h(s, a, o) + (1 - P_h(s, a, o)) V^{(t)}(s) \Big) \pi(s, a)$$

which we solve with the following linear program

Maximize: $V(s)$
Subject to: $\sum_a \pi(s, a) = 1$
$\forall a \in \mathcal{A}, \ \pi(s, a) \geq 0$
$\forall o \in \mathcal{O}, \ V(s) \leq \sum_a (P_h(s, a, o) \mathcal{R}_h(s, a, o)$
$+ (1 - P_h(s, a, o)) V^{(t)}(s)) \pi(s, a)$

(3)

**Theorem 5.2** *The operator $T_h$ for the opt-out game is a non-expansion in any weighted $L_2$ norm.*

**Proof:** Consider two value functions, $V_1$ and $V_2$, with $\|V_1 - V_2\|_\infty = \epsilon$. We show that $T_h$ is pointwise non-expansive, which implies that it is a non-expansion in any weighted $L_2$ norm by Lemma 5.1. We will write the coefficients of the linear program constraint for $T_h$ with opponent action $k$ and value function $V_1$ as $O_1^k$ and the corresponding constraint for value function $V_2$ as $O_2^k$. The corresponding constraints can then be expressed in terms of a dot product, e.g., for $V_1(s)$ we get $V(s) \leq O_1^k \cdot \pi(s)$, where $\pi(s)$ is a vector of probabilities for each action.

Assume $|V_1(s) - V_2(s)| = \epsilon$. This implies $\|O_2^k - O_2^k\|_\infty \leq \epsilon, \forall k$. We will call $v_1$ the minimax solution for $T_h V_1$ (with policy $\pi_1$) and call $v_2$ the minimax solution for $T_h V_2$ (with policy $\pi_2$). We must show $|v_1 - v_2| \leq \epsilon$. In the following, we will use the fact that there must exist at least one tight constraint on $V$ and we define $j$ to be the minimizing opponent action in the second summation in the second line below. We also assume, without loss of generality, $v_1 > v_2$.

$$\begin{aligned}
v_1 - v_2 &= \min_k O_1^k \cdot \pi_1 - \min_k O_2^k \cdot \pi_2 \\
&\leq \min_k O_1^k \cdot \pi_1 - \min_k O_2^k \cdot \pi_1 \\
&\leq \min_k O_1^k \cdot \pi_1 - O_2^j \cdot \pi_1 \\
&\leq O_1^j \cdot \pi_1 - O_2^j \cdot \pi_1 \\
&\leq (O_1^j - O_2^j) \cdot \pi_1 \\
&\leq \epsilon. \ \blacksquare
\end{aligned}$$

---

[5] The opt-out game described here is different from a *Quitting Game* (Solan & Vieille, 2001) in several ways. It is more limited in that it is just two-player and zero-sum, not n-player and general sum. It is more general, in that that both players are moving along a Markov chain in which different payoff matrices may be associated with each state.

The error bounds and convergence results for the optimal stopping version of linear TD and LSTD all generalize to the opt-out game as immediate corollaries of this result.[6]

## 6 Efficient Learning in Markov Games

The previous section described some specific cases where value function approximation is convergent for Markov games. As with MDPs, we may be tempted to use a learning method with weaker guarantees to attack large, generic Markov games. In this section, we demonstrate good performance using value function approximation for Markov games. While the stronger convergence results of the previous section do not apply here, the general results of Section 3 do, as well as the generic approximation policy iteration bounds from (Bertsekas & Tsitsiklis, 1996).

A limitation of $Q$-learning with function approximation is that it often requires many training samples. In the case of Markov games this would require solving a large number of linear programs. This would lead to slow learning, which would slow down the iterative process by which the designer of the value function approximation architecture selects features (basis functions). In contrast to $Q$-learning, LSPI (Lagoudakis & Parr, 2001) is a learning algorithm that makes very efficient use of data and converges faster than conventional methods. This makes LSPI well-suited to value function approximation in Markov games as it has the potential of reducing the number of linear programs required to be solved. We briefly review the LSPI framework and describe how it extends to Markov games.

LSPI is a model-free approximate policy iteration algorithm that learns a good policy for any MDP from a corpus of stored samples taken from that MDP. LSPI works with $Q$-functions instead of $V$-functions. The $Q_\pi$ function for a policy $\pi$ is approximated by a linear combination of basis functions $\phi_i$ (features) defined over states *and* actions:

$$\widehat{Q}_\pi(s,a,w) = \sum_{i=1}^{k} \phi_i(s,a)w_i = \phi(s,a)^\mathsf{T} w \ .$$

Let $\Phi$ be a matrix with the basis functions evaluated at all points similar to the one in the previous section, but now $\Phi$ is of size $(|\mathcal{S}||\mathcal{A}| \times k)$. If we knew the transition model and reward function, we could, in principle, compute the weights $w^\pi$ of $\widehat{Q}_\pi$ by defining and solving the system $\mathbf{A}w^\pi = b$, where $\mathbf{A} = \Phi^\mathsf{T}(\Phi - \gamma \mathbf{P}^\pi \Phi)$ and $b = \Phi^\mathsf{T}\mathcal{R}$.

Since there is no model available in the case of learning, LSPI uses sample experiences from the environment in place of $\mathbf{P}^\pi$ and $\mathcal{R}$. Given a set of samples, $D = \{(s_{d_i}, a_{d_i}, s'_{d_i}, r_{d_i}) \mid i = 1, 2, \ldots, L\}$, LSPI constructs ap-

---

[6]An alternative proof of this theorem can be obtained via the *conservative non-expansion property* of the minimax operator, as defined in Szepesvari and Littman (1999).

proximate versions of $\Phi$, $\mathbf{P}^\pi \Phi$, and $\mathcal{R}$ as follows:

$$\widehat{\Phi} = \begin{pmatrix} \phi(s_{d_1}, a_{d_1})^\mathsf{T} \\ \cdots \\ \phi(s_{d_i}, a_{d_i})^\mathsf{T} \\ \cdots \\ \phi(s_{d_L}, a_{d_L})^\mathsf{T} \end{pmatrix} \quad \widehat{\mathbf{P}^\pi \Phi} = \begin{pmatrix} \phi(s'_{d_1}, \pi(s'_{d_1}))^\mathsf{T} \\ \cdots \\ \phi(s'_{d_i}, \pi(s'_{d_i}))^\mathsf{T} \\ \cdots \\ \phi(s'_{d_L}, \pi(s'_{d_L}))^\mathsf{T} \end{pmatrix}$$

$$\widehat{\mathcal{R}} = \begin{pmatrix} r_{d_1} \\ \cdots \\ r_{d_i} \\ \cdots \\ r_{d_L} \end{pmatrix}$$

Assume that initially $\widehat{\mathbf{A}} = 0$ and $\widehat{b} = 0$. For a fixed policy $\pi$, a new sample $(s, a, r, s')$ contributes to the approximation according to the following update equations:

$$\widehat{\mathbf{A}} \leftarrow \widehat{\mathbf{A}} + \phi(s,a)\Big(\phi(s,a) - \gamma\phi(s', \pi(s'))\Big)^\mathsf{T} \ ,$$

$$\widehat{b} \leftarrow \widehat{b} + \phi(s,a)r \ .$$

The weights are computed by solving $\widehat{\mathbf{A}} w^\pi = \widehat{b}$. This approach is similar to the LSTD algorithm (Bradtke & Barto, 1996). Unlike LSTD, which defines a system of equations relating $V$-values, LSPI is defined over $Q$-values. Each iteration of LSPI yields the $Q$-values for the current policy $\pi^{(t)}$, implicitly defining the next policy $\pi^{(t+1)}$ for policy iteration as the greedy policy over the current $Q$-function:

$$\pi^{(t+1)}(s) = \arg\max_{a \in \mathcal{A}} \widehat{Q}_{\pi^{(t)}}(s,a) = \arg\max_{a \in \mathcal{A}} \phi(s,a)^\mathsf{T} w^{\pi^{(t)}} .$$

Policies and $Q$-functions are never represented explicitly, but only implicitly through the current set of weights, and they are computed on demand.

A key feature of LSPI is that it can reuse the same set of samples even as the policy changes. For example, suppose the corpus contains a transition from state $s$ to state $s'$ under action $a_1$ and $\pi_i(s') = a_2$. This is entered into the $\widehat{\mathbf{A}}$ matrix as if a transition were made from $(s, a_1)$ to $(s', a_2)$. If $\pi_{i+1}(s')$ changes the action for $s'$ from $a_2$ to $a_3$, then the next iteration of LSPI enters a transition from $(s, a_1)$ to $(s', a_3)$ into the $\widehat{\mathbf{A}}$ matrix. Sample reuse is possible because the dynamics for state $s$ under action $a_1$ have not changed.

The generalization of LSPI to Markov games is quite straightforward. Instead of learning $\widehat{Q}(s,a)$ functions, we learn $\widehat{Q}(s,a,o)$ functions defined as $\widehat{Q}(s,a,o) = \phi(s,a,o)^\mathsf{T} w$. The minimax policy $\pi$ at any given state $s$ is determined by

$$\pi(s) = \arg\max_{\pi(s) \in \Omega(\mathcal{A})} \min_{o \in \mathcal{O}} \sum_{a \in \mathcal{A}} \pi(s,a) \phi(s,a,o)^\mathsf{T} w \ ,$$

and computed by the solving the following linear program:

Maximize: $V(s)$
Subject to: $\sum_{a \in \mathcal{A}} \pi(s,a) = 1$
$\forall a \in \mathcal{A}, \ \pi(s,a) \geq 0$
$\forall o \in \mathcal{O}, \ V(s) \leq \sum_{a \in \mathcal{A}} \pi(s,a)\phi(s,a,o)^\mathsf{T} w$

Finally, the update equations have to be modified to account for the opponent's action and the distribution over next actions since the minimax policy is, in general, stochastic:

$$\widehat{\mathbf{A}} \leftarrow \widehat{\mathbf{A}} + \phi(s,a,o)\Big(\phi(s,a,o) - \gamma \sum_{a' \in \mathcal{A}} \pi(s',a')\phi(s',a',o')\Big)^{\mathsf{T}},$$

$$\widehat{b} \leftarrow \widehat{b} + \phi(s,a,o)r,$$

for any sample $(s, a, o, r, s')$. The action $o'$ is the minimizing opponent's action in computing $\pi(s')$.

## 7 Experimental Results

We tested our method on two domains that can be modeled as zero-sum Markov games: a simplified soccer game, and a router/server flow control problem. Soccer, by its nature is a competitive game, whereas the flow control problem would normally be a single agent problem (either for the router or the server). However, viewing the router and the server as two competitors (although they may not really be) can yield solutions that handle worst-case scenarios for both the server and the router.

### 7.1 Two-Player Soccer Game

This game is a simplification of soccer (Littman, 1994) played by two players on a rectangular grid board of size $R \times C$. Each player occupies one cell at each time; the players cannot be in the same cell at the same time. The state of the game consists of the positions of the two players and the possession of the ball. Initially the players are placed randomly in the first and the last column respectively[7] and the ball is given to either player randomly. There are 5 actions for each player : up (U), down (D), left (L), right (R), and stand (S). At each time step the players decide on which actions they are going to take and then a fair coin is flipped to determine which player moves first. The players move one at a time in the order determined by the coin flip. If there is a collision with a border or with another player during a move, the player remains in its current position. If the player with the ball (attacker) runs into the opponent (defender), the ball is passed to the opponent. Therefore, the only way for the defender to steal the ball is to be in the square into which the attacker intends to move. The attacker can cross the goal line and score into the defender's goal, however the players cannot score into their own goals. Scoring for the agent results in an immediate reward of $+1$, whereas scoring for the opponent results in an immediate reward of $-1$, and the game ends in either case. The discount factor for the problem is set to a value less than 1 to encourage early scoring.

---
[7]We decided to make this modification in the original game (Littman, 1994), because good players will never leave the central horizontal band between the two goals (the goal zone) if they are initialized within it, rendering the rest of the board useless.

The discrete and nonlinear dynamics of the game make it particularly difficult for function approximation. We experimented with a wide variety of basis functions, but ultimately settled on a fairly simple, model-free set that yielded reasonable policies. We believe that we could do better with further experimentation, but these simple basis functions are sufficient to demonstrate the effectiveness of value function approximation.

The basic block of our set consists of 36 basis functions. This block is replicated for each of the 25 action pairs; each block is active only for the corresponding pair of actions, so that each pair has its own weights. Thus, the total number of basis functions, and therefore the number of free parameters in our approximation, is 900. Note that the exact solution for a $4 \times 4$ grid requires $12,000$ $Q$-values, whereas for a $40 \times 40$ grid requires $127,920,000$ $Q$-values. It is clear that large grids cannot be solved without relying on some sort of function approximation.

The 36 functions in the basic block are not always active for a fixed pair of actions. They are further partitioned to distinguish between 4 cases according to the following binary propositions:

$P_1$ "*The attacker is closer to the defender's goal than the defender*", i.e, there is path for the attacker that permits scoring without interception by the defender.

$P_2$ "*The defender is close to the attacker*", which means that the defender is within a Manhattan distance of 2 or, in other words, that there is at least one pair of actions that might lead to a collision of the two players.

The validity of the propositions above can be easily detected given any state of the game. The basis functions used in each of the 4 cases are as follows:

- $P_1$ and not $P_2$
    - $1.0$ : a constant term
    - $D_H P_A G_D$ : the horizontal distance[8] of the attacker from the defender's goal
    - $SD_V P_A G$ : the signed vertical distance of the attacker from the goal zone
    - $D_H P_A G_D \times SD_V P_A G$

- $P_1$ and $P_2$
    - $1.0, D_H P_A G_D, SD_V P_A G, D_H P_A G_D \times SD_V P_A G$
    - $P_A UG$ : indicator that the attacker is at the upper end of the goal zone
    - $P_A LG$ : indicator that the attacker is at the lower end of the goal zone
    - $SD_H P_A P_D$ : the signed horizontal distance between the two players
    - $SD_V P_A P_D$ : the signed vertical distance between the two players

- not $P_1$ and not $P_2$
    - same as above, excluding $SD_H P_A P_D$

---
[8]All distances are Manhattan distances scaled to $[0, 1]$.

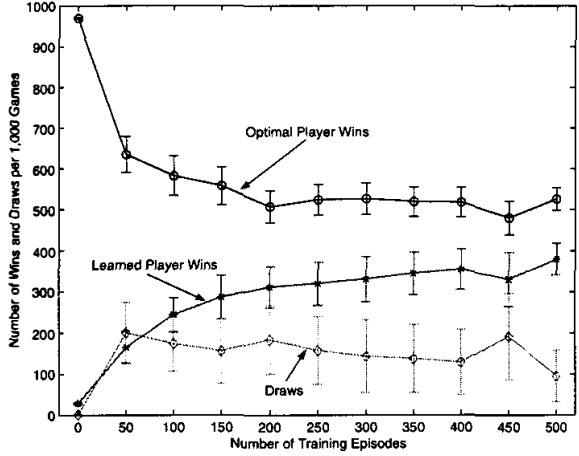

Figure 1: Number of Wins and Draws for Each Player

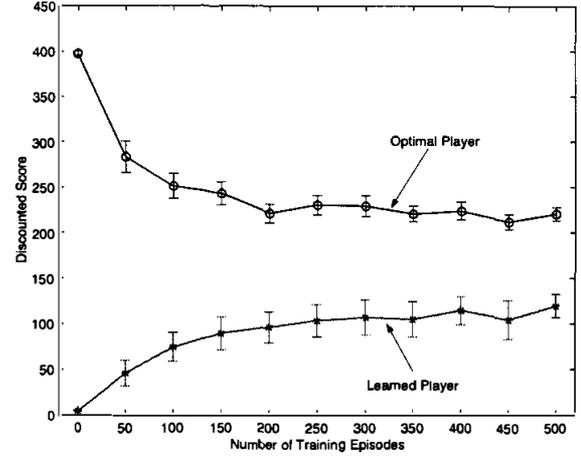

Figure 2: Total (discounted) Score for Each Player ($\gamma = 0.9$)

- not $P_1$ and $P_2$
  - $D_H P_A G_D$, $SD_V P_A G$, $D_H P_A G_D \times SD_V P_A G$, $P_A UG$, $P_A LG$
  - $P_D G_D L$ : indicator that the defender is by the goal line of his goal
  - $P_D WG$: indicator that the defender is within the goal zone
  - 10 indicators for the 10 possible positions[9] of the defender within Manhattan distance of 2

Notice that all basis functions are expressed in the relative terms of *attacker* and *defender*. In our experiments, we always learned a value function from the perspective of the attacker. Since the game is zero-sum, the defender's weights and value function are simply the negation of the attacker's. This convenient property allowed to learn from one agent's perspective but use the same value function for either offense or defense during a game.

We conducted a series of experiments on a 4 × 4 grid to compare our learned player with the optimal minimax player (obtained by exact policy iteration) for this grid size. Training samples were collected by "random games", i.e., complete game trajectories where the players take random actions independently until scoring occurs. The average length of such a game in the 4 × 4 grid is about 60 steps. We varied the number of games from 0 to 500 in increments of 50. In all cases, our training sets had no more than 40, 000 samples. For each data set, LSPI was run until convergence or until a maximum of 25 iterations was reached. The final policy was compared against the optimal player in a tournament of 1, 000 games. A game was called a draw if scoring did not occur within 100 steps. The entire experiment was repeated 20 times and the results were averaged.

The results are summarized in Figures 1 - 2, with error bars indicating 95% confidence intervals. The player corresponding to 0 training episodes is a purely random player that selects actions with uniform probability. Clearly, LSPI finds better players with more training data. However, function approximation errors may prevent it from ultimately reaching the performance of the optimal player.

We also applied our method to bigger soccer grids where obtaining the exact solution or even learning with a conventional method is prohibitively expensive. For these grids, we used a larger set of basis functions (1400 total) that included several crossterms compared to the set described above. These terms were not included in the 4 × 4 case because of linear dependencies caused due to the small grid size. Below are listed the extra basis functions used *in addition* to the original ones listed above:

- $P_1$ and not $P_2$
  - $(D_H P_A G_D)^2$, $(SD_V P_A G)^2$
  - $P_A UG$, $P_A LG$, $P_A WG$
- $P_1$ and $P_2$
  - $(D_H P_A G_D)^2$, $(SD_V P_A G)^2$, $P_A WG$
  - $SD_H P_A P_D \times SD_V P_A P_D$
- not $P_1$ and not $P_2$
  - $(D_H P_A G_D)^2$, $(SD_V P_A G)^2$, $P_A WG$, $SD_V P_A P_D$
  - $(SD_V P_A P_D)^2$, $(SD_H P_A P_D)^2$
  - $SD_H P_A P_D \times SD_V P_A P_D$, $P_D WG$
- not $P_1$ and $P_2$
  - $(D_H P_A G_D)^2$, $(SD_V P_A G)^2$, $P_A WG$

We obtained a benchmark player for an 8 × 8 grid by generating 2 samples, one for each ordering of player moves for each of the 201, 600 states and action combinations, and by using these artificially generated samples to do a uniform projection into the column space of $\Phi$. In some sense, this benchmark player is the best player we could hope to achieve for the set of basis functions under consideration, assuming that uniform projection is the best one. Our anecdotal evaluation of this player is that it was very strong.

---

[9] There are actually 12 such positions, but two of them (the ones behind the attacker) are excluded because if the defender was there, $P_1$ would be true.

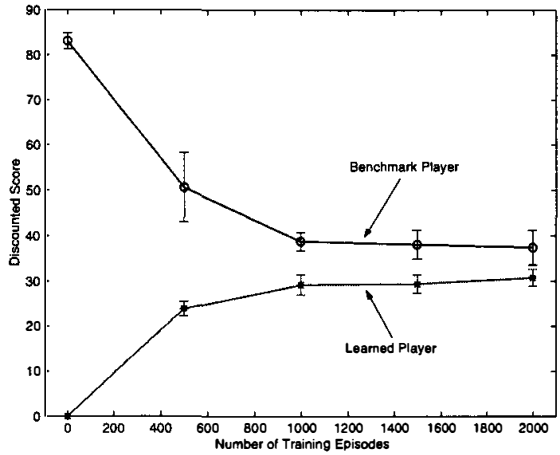
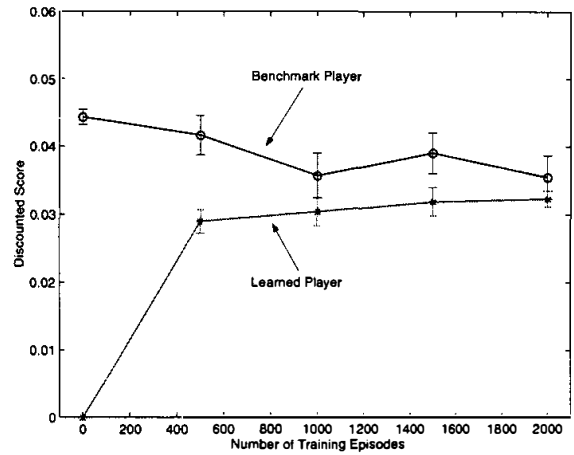

Figure 3: 8 × 8 Soccer: Discounted Total Score, $\gamma = 0.8$

Figure 4: 40 × 40 Soccer: Discounted Total Score, $\gamma = 0.8$

We conducted a series of experiments against learned players obtained using training sets of different sizes. Training samples were again collected by "random games" with a maximum length of 1,000 steps. The average length was about 180 steps. We varied the number of such games from 0 to 2,000 in increments of 500. In all cases, our training sets had no more than 400,000 samples. For each data set, LSPI was run until convergence, typically about 8 iterations. The final policy was compared against the benchmark player in a tournament of 1,000 games. A game was called a draw if scoring had not occurred within 300 steps. Average results over 10 repetitions of the experiment are shown in Figure 3. Again, LSPI produces a strong player after a relatively small number of games.

To demonstrate generalization abilities we took all policies learned in the 8 × 8 grid and we tested them in a 40 × 40 grid. Our anecdotal evaluation of the benchmark player in the 40 × 40 grid was that played nearly as well as in the 8 × 8 grid. The learned policies also performed quite reasonably and almost matched the performance of the benchmark player as shown in Figure 4. Note that we are comparing policies trained in the 8 × 8 grid. We never trained a player directly for the 40 × 40 grid since random walks in this domain would result mostly in random wandering with infrequent interaction between players or scoring, requiring a huge number of samples.

### 7.2 Flow Control

In the *router/server flow control problem* (Altman, 1994), a router is trying to control the flow of jobs into a server buffer under unknown, and possibly changing, service conditions. This problem can be modeled as an MDP with the server being an uncertain part of the environment. However, to provide worst-case guarantees the router can view the server as an opponent that plays against it. This viewpoint enables to router to adopt control policies that perform well under worst-case/changing service conditions.

The state of the game is the current length of the buffer. The router can choose among two actions, low (L) and high (H), corresponding to a low ($PA_L$) and a high ($PA_H$) probability of a job arrival to the buffer at the current time step, with $0 < PA_L < PA_H \le 1$. Similarly, the server can choose among low (L) and high (H), corresponding to a low ($PD_L$) and a high ($PD_H$) probability of a job departure from the buffer at the current time step, with $0 \le PD_L < PD_H < 1$. Once the agents pick their actions, the size of the buffer is adjusted to a new state according to the chosen probabilities and the game continues. The immediate cost $\mathcal{R}(s, a, o)$ for each transition depends on the current state $s$ and the actions $a$ and $o$ of the agents:

$$\mathcal{R}(s, a, o) = c(s) + \alpha \times PA_a + \beta \times PD_o,$$

where $c(s)$ is a real non-decreasing convex function, $\alpha \le 0$, and $\beta \ge 0$. $c(s)$ is related to the holding cost per time step in the buffer, $\alpha$ is related to the reward for each incoming job, and $\beta$ is related to the cost for the quality of service. The router attempts to minimize the expected discounted cost, whereas the server strives to maximize it. The discount factor is set to 0.95.

Under these conditions, the optimal policies can be shown to have an interesting threshold structure, with mostly deterministic choices and randomization in at most one state (Altman, 1994). However, the exact thresholds and probabilities cannot be easily determined analytically from the parameters of the problems which might actually be unknown. These facts make the problem suitable for learning with function approximation. We used a polynomial of degree 3 ($1.0, s, s^2, s^3$) as the basic block of basis functions to approximate the $Q$ function for each pair of actions. That resulted in a total of 16 basis functions.

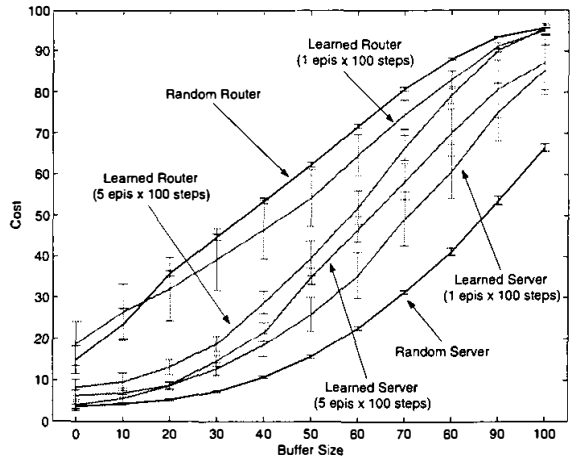

Figure 5: Flow Control: Performance Against the Optimal.

We tested our method on a buffer of length 100 with: $PA_L = 0.2$, $PA_H = 0.9$, $PD_L = 0.1$, $PD_H = 0.8$, $c(s) = 10^{-4}s^2$, $\alpha = -0.1$, $\beta = +1.5$. With these settings neither a full nor an empty buffer is desirable for either player. Increasing the buffer size beyond 100 does not cause any change in the final policies, but will require more training data to cover the critical area (0-100).

Training samples were collected as random episodes of length 100 starting at random positions in the buffer. Figure 5 shows the performance of routers and servers learned using 0 (random player), 1, or 5 training episodes (the error bars represent the 95% confidence intervals). Routers are tested against the optimal server and servers are tested against the optimal router.[10] Recall that the router tries to minimize cost whereas the server is trying to maximize. With about 100 or more training episodes the learned player is indistinguishable from the optimal player.

## 8 Conclusions and Future Work

We have demonstrated a framework for value function approximation in zero-sum Markov games and we have generalized error bounds from MDPs to the Markov game case, including a variation on the optimal stopping problem. A variety of approximate MDP algorithms can be generalized to the Markov game case and we have explored a generalization of the least squares policy iteration (LSPI) algorithm. This algorithm was used to find good policies for soccer games of varying sizes and was shown to have good generalization ability: policies trained on small soccer grids could be applied successfully to larger grids. We also demonstrated good performance on a modest flow control problem.

One of the difficulties with large games is that it is difficult to define a gold standard for comparison since evaluation requires an opponent, and optimal adversaries are not known for many domains. Nevertheless, we hope to attack larger problems in our future work, particularly the rich variety of flow control problems that can be addressed with the Markov game framework. Queuing problems have been carefully studied and there should be many quite challenging heuristic adversaries available. We also plan to consider team-based games and general sum games.

**Acknowledgements** We are grateful to Carlos Guestrin and Michael Littman for helpful discussions. Michail G. Lagoudakis was partially supported by the Lilian Boudouri Foundation.

---

[10]The "optimal" behavior for this problem was determined by using a tabular representation and an approximate model obtained by generating 1000 next state samples for each $Q$-value.